\documentclass[11pt]{article}

\usepackage[]{EMNLP2023}
\usepackage{times}
\usepackage{latexsym}

\usepackage[T1]{fontenc}
\usepackage[utf8]{inputenc}

\usepackage{microtype}

\usepackage{inconsolata}

\usepackage{times}
\usepackage{soul}
\usepackage{url}
\usepackage{caption}
\usepackage{graphicx}
\usepackage{amsmath}
\usepackage{amsthm}
\usepackage{booktabs}
\usepackage{algorithm}
\usepackage{algorithmic}
\usepackage[switch]{lineno}
\usepackage{multirow}
\usepackage{makecell}  % 支持单元格换行
\usepackage{amsfonts}
\usepackage{forest}
\usepackage{hyperref}
\newtheorem{definition}{Definition} % 定义 "definition" 环境

\usepackage{tikz-qtree}
\usetikzlibrary{trees,decorations.text,math,calc, positioning, arrows.meta}
\usepackage{xcolor}
\definecolor{Red}{RGB}{190,20,42}
\definecolor{CY}{RGB}{233,246,254}
\usepackage[T1]{fontenc}
\usepackage[utf8]{inputenc} % 兼容 UTF-8
\usepackage{lmodern}        % 改善默认字体
\usepackage{mathptmx}  % 或者 \usepackage{newtxtext,newtxmath}
\title{A Survey of Link Prediction in N-ary Knowledge Graphs}

\author{Jiyao Wei, Saiping Guan\thanks{Corresponding Authors.}, Da Li, Xiaolong Jin$^{*}$, Jiafeng Guo, Xueqi Cheng\\
School of Computer Science and Technology, University of Chinese Academy of Sciences; \\
Key Laboratory of Network Data Science and Technology, \\
Institute of Computing Technology, Chinese Academy of Sciences. \\
  \texttt{\{weijiyao20z, guansaiping, lida, jinxiaolong, guojiafeng, cxq\}@ict.ac.cn}}

\begin{document}
\maketitle
\begin{abstract}
  N-ary Knowledge Graphs (NKGs) are a specialized type of knowledge graph designed to efficiently represent complex real-world facts. Unlike traditional knowledge graphs, where a fact typically involves two entities, NKGs can capture n-ary facts containing more than two entities. Link prediction in NKGs aims to predict missing elements within these n-ary facts, which is essential for completing NKGs and improving the performance of downstream applications. This task has recently gained significant attention. In this paper, we present the first comprehensive survey of link prediction in NKGs, providing an overview of the field, systematically categorizing existing methods, and analyzing their performance and application scenarios. We also outline promising directions for future research.
\end{abstract}

\section{Introduction}
Since Google introduced Knowledge Graph (KG) to enhance its search services, KGs have attracted growing attention from both academia and industry~\cite{lehmann2015dbpedia}. A traditional KG stores numerous facts, typically represented in the form of triples ($h$, $r$, $t$), indicating a specific relation $r$ between a head entity $h$ and a tail entity $t$, such as ($Biden$, $the$ $President$ $of$, $the$ $USA$). 

However, many real-world facts involve more than two entities, requiring a more expressive representation. N-ary Knowledge Graphs (NKGs) address this need by enabling the representation of complex facts involving multiple entities, commonly referred to as n-ary facts. For instance, the fact \textit{``Einstein studied physics at the University of Zurich and received his PhD"} can be represented as \{$person: Einstein,$ $institution: Uni.\;Zurich,$ $major: Physics,$ $degree: PhD$\} in NKGs. 

N-ary facts are prevalent in real-world scenarios~\cite{fatemi2019knowledge}. Statistically, in Freebase, over a third of entities are involved in n-ary facts~\cite{wen2016representation}, and more than 61\% facts are n-ary facts~\cite{fatemi2019knowledge}. Like traditional KGs, NKGs are inevitably incomplete, due to the complex process of their construction~\cite{li2024hje}. The incompletion of NKGs hinders the performance of downstream applications, including information retrieval~\cite{zhao2020research} and recommendation systems~\cite{liang2023graph}. To address this, link prediction in NKGs is proposed to predict missing elements in facts therein, helping populate and enrich NKGs~\cite{wen2016representation}.

Traditional link prediction methods for KGs encode facts as triples. To handle n-ary facts, they decompose each n-ary fact into multiple triples, such as introducing Compound Value Type (CVT) entities in Freebase~\cite{bollacker2008freebase}. However, this decomposition complicates inference, leads to structural information loss, increases model parameters, and risks incorrect reasoning~\cite{wen2016representation}. More details on the disadvantages of the decomposition are shown in Appendix~\ref{decomposition}.

As shown in Figure~\ref{fig_nkg_articles}, recent efforts have increasingly focused on directly modeling n-ary facts without decomposition. Methods for link prediction in NKGs fall into three categories: spatial mapping-based~\cite{wen2016representation}, tensor decomposition-based, and neural network-based approaches~\cite{guan2019link}. These methods address both general scenarios~\cite{wen2016representation} and specialized scenarios, including temporal~\cite{hou2023temporal}, few-shot~\cite{zhang2022true}, and inductive settings~\cite{ali2021improving}.

Although there exist numerous surveys on link prediction in KGs, covering general~\cite{wang2017knowledge,guan2022event}, temporal~\cite{cai2022temporal,wang2023survey}, multi-modal KGs~\cite{zhu2022multi,peng2023multi}, and sparse KGs~\cite{chen2023generalizing}, none specifically focus on NKGs. More statistics of existing surveys are shown in Appendix~\ref{survey}. Despite the rapid development of link prediction methods for NKGs, with nearly 50 methods proposed, existing surveys only briefly mention it, lacking a comprehensive and in-depth analysis. A dedicated survey on link prediction in NKGs is crucial for understanding the progress, challenges, and future directions of this field. This paper aims to fill this gap by providing a detailed and timely survey on link prediction in NKGs, facilitating further research in this area.

\begin{figure}[t]
  \centering
  \includegraphics[width=0.47\textwidth]{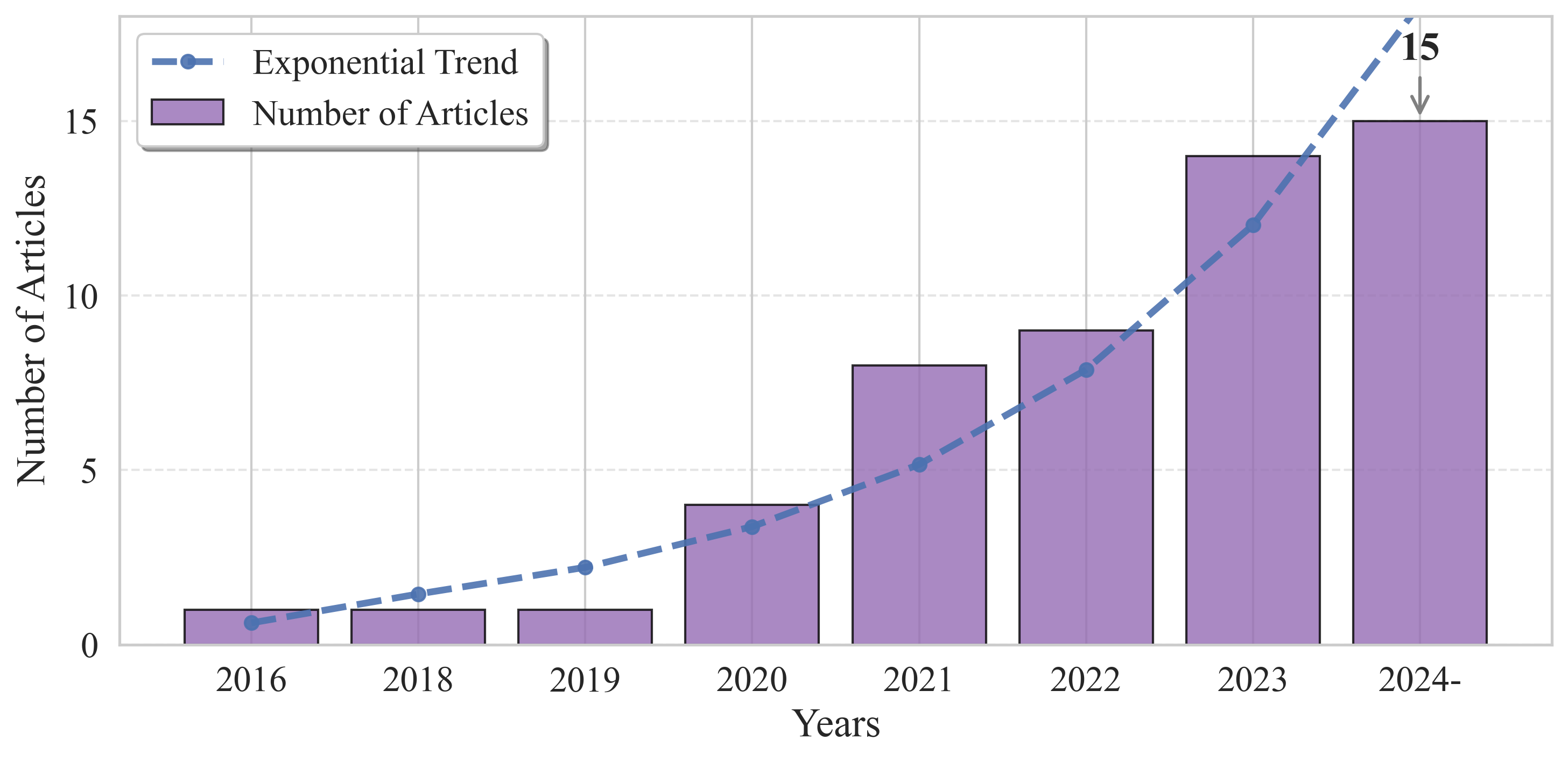}
  \caption{Number of articles published each year (2016-) on link prediction in NKGs.}
  \label{fig_nkg_articles}
\end{figure}

The rest of this paper is organized as follows: Section~\ref{section2} introduces the related definitions of link prediction in NKGs and categorizes existing methods. Section~\ref{section3} analyzes existing methods. Section~\ref{experiments} reports their performance on benchmarks. Section~\ref{section5} highlights representative applications. Finally, Section~\ref{section6} suggests future research directions. Compilation and details of papers used for the survey can be found via our repository\footnote{\url{https://github.com/JiyaoWei/LP_NKGs}}.

\section{\textbf{Preliminary}}\label{section2}
In this section, we introduce the related definitions of link prediction in NKGs, the formalizations of n-ary facts, the classifications of existing methods for link prediction in NKGs, and the applicability of link prediction in KGs and NKGs.

\subsection{\textbf{Definition}}

\begin{definition}
  \textbf{KG}:
  a set of facts, each of which is represented as a triple $(h,r,t)$, where $h$ and $t$ denote its head entity and tail entity, respectively, and $r$ denotes the relation between them.
  \end{definition}
  
  For example, in a KG, the fact \textit{``Biden is the president of the USA"} is represented as $(Biden$, $the$ $President$ $of$, $the$ $USA)$.

  \begin{definition}
  \textbf{NKG}:
  a set of facts, each of which may contain more than two entities, which is also referred to as an n-ary fact. 
  \end{definition}
  
  For example, in an NKG, the n-ary fact \textit{``Einstein studied physics at the University of Zurich and received his PhD."} contains four entities, including \textit{Einstein, the University of Zurich, Physics}, and \textit{PhD}. Therefore, it is called a 4-ary fact. 
  
  \begin{definition}
  \textbf{Link prediction in NKGs}:
  predict missing elements in facts in NKGs based on the existing facts. 
  \end{definition}
  
  For example, predict the missing entity \textit{the University of Zurich} in the n-ary fact \textit{``Einstein studied physics at ? and received his PhD"}.

  \subsection{\textbf{The formalizations of N-ary Facts\label{repre_nary}}}
  Typical n-ary fact formalizations include hyperedge, role-value pair, and hyper-relation formalizations, as illustrated in Figure~\ref{fig_1}.
  
  \subsubsection{Hyperedge Formalization} 
\begin{figure}[t]
      \centering
      \includegraphics[width=0.49\textwidth]{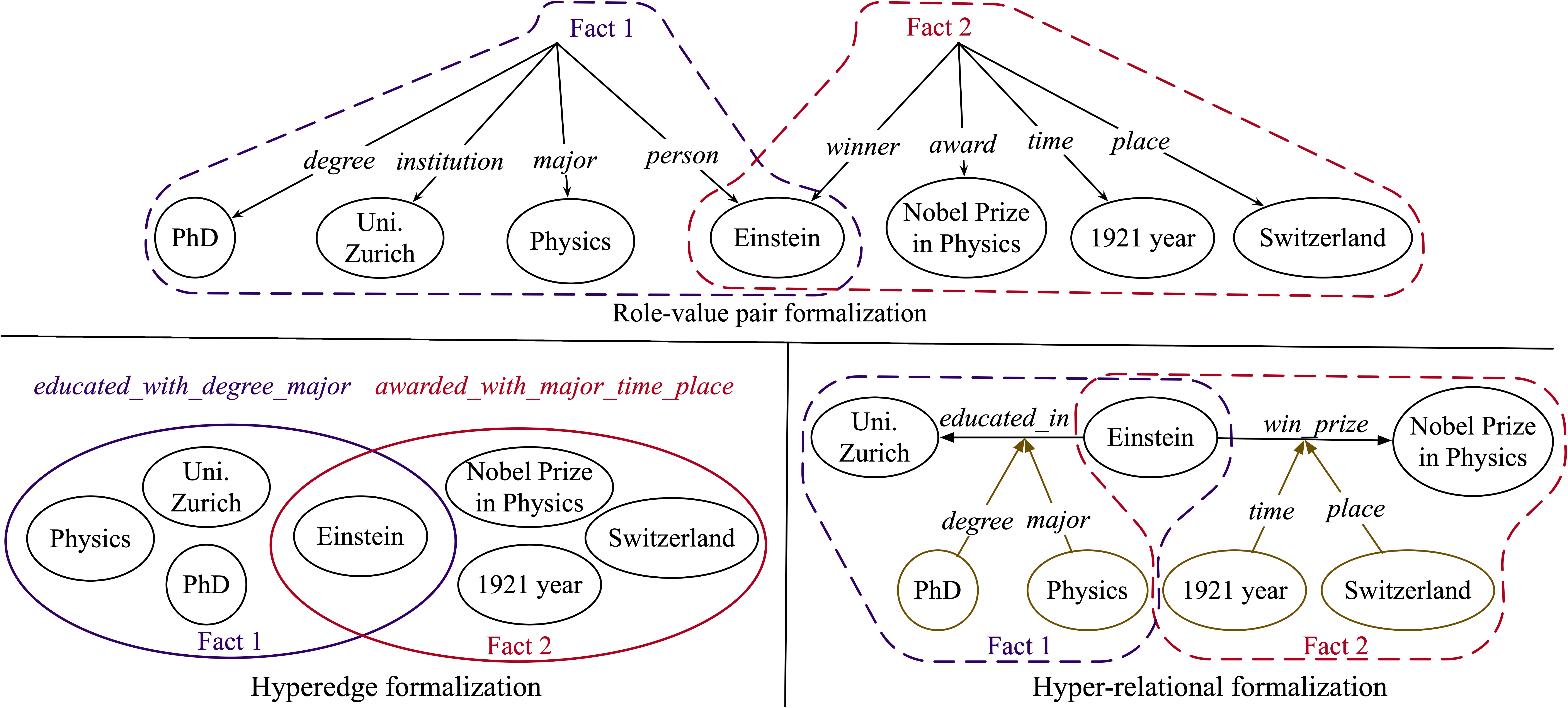}
      \caption{Examples of different formalizations of n-ary facts.}
      \label{fig_1}
  \end{figure}

  A hyperedge connects all entities in an n-ary fact~\cite{wen2016representation}, e.g., $(H, e_1, ..., e_n)$, where $e_*$ is the *-th entity and hyperedge $H$ indicates the role of each entity in the fact.
  For example, the facts in Figure~\ref{fig_1} can be represented as follows:
  
  Fact 1: $(educated\_with\_degree\_major,$ $Einstein,$ $Uni.$ $Zurich,$ $PhD,$ $Physics)$. 
  
  Fact 2: $(awarded\_with\_time\_place,$ $Einstein,$ $Nobel$ $Prize$ $in$ $Physics,$ $Switerland$, $1921$ $year)$. 
  
  Note that under the hyperedge formalization, the entities in the formalized fact are ordered with a fixed number of entities. Each position in the hyperedge represents a fixed role. 
  The hyperedge formalization directly builds connections between multiple entities in an n-ary fact. When dealing with binary facts, this formalization is simplified to the triple formalization in traditional KGs. 

  \subsubsection{Role-value Pair Formalization}
  An n-ary fact is formulated as multiple role-value pairs~\cite{guan2019link}, such as $\{r_i:v_i\}_{i=1}^n$, where value $v_i$ is an entity and plays the role $r_i$ in the fact; $n$ is the total number of entities within the fact. For example, the two facts about Einstein introduced above can be represented as follows:
  
  Fact 1: $\{person: Einstein,$

  \;\;\;\;\;\;\;\;\;\;\;\;$institution: Uni.\;Zurich, $
  
  \;\;\;\;\;\;\;\;\;\;\;\;$degree: PhD, $

  \;\;\;\;\;\;\;\;\;\;\;\;$major: Physics\}.$

  Fact 2: $\{winner: Einstein,$
  
  \;\;\;\;\;\;\;\;\;\;\;\;$award: Nobel\;Prize\;in\;Physics,$
  
  \;\;\;\;\;\;\;\;\;\;\;\;$place: Switerland,$

  \;\;\;\;\;\;\;\;\;\;\;\;$time: 1921\;year\}.$
  
  Note that in the role-value pair formalization, the role-value pairs within a fact are unordered and may involve an arbitrary number of entities. This representation offers flexibility in specifying the roles of entities in n-ary facts. However, it fails to account for the varying importance or prominence of different entities within the same fact.

  \subsubsection{Hyper-relational Formalization}
  An n-ary fact is formulated as a primary triple coupled with a set of qualifier role-value pairs~\cite{rosso2020beyond,guan2020neuinfer}, i.e., $((h, r, t), \{r_i:v_i\}_{i=1}^{n-2})$, where $h$ and $t$ are the head entity and tail entity of the fact and $r$ denotes the relation between them; role-value pairs $\{r_i:v_i\}_{i=1}^{n-2}$ qualify the triple $(h, r, t)$. For example, the two facts about Einstein introduced above can be represented as follows:
  
  Fact 1: $((Einstein, educated, Uni.\; Zurich), $

  \;\;\;\;\;\;\;\;\;\;\;\;$-|\{major: Physics, $
  
  \;\;\;\;\;\;\;\;\;\;\;\;$-|degree: PhD\})$.
  
  Fact 2: $((Einstein, won, Nobel\;Prize\;in$ $\;Physics), $
  
  \;\;\;\;\;\;\;\;\;\;\;\;$-|\{time: 1921$ $year, $
  
  \;\;\;\;\;\;\;\;\;\;\;\;$-|place: Switerland\})$.
  
  Note that when there is no clear subject (i.e., head entity) or object (i.e., tail entity) in the facts, it is not appropriate to use the hyper-relational formalization. Additionally, no matter which formalization is used, link prediction in NKGs aims to predict missing elements in facts.

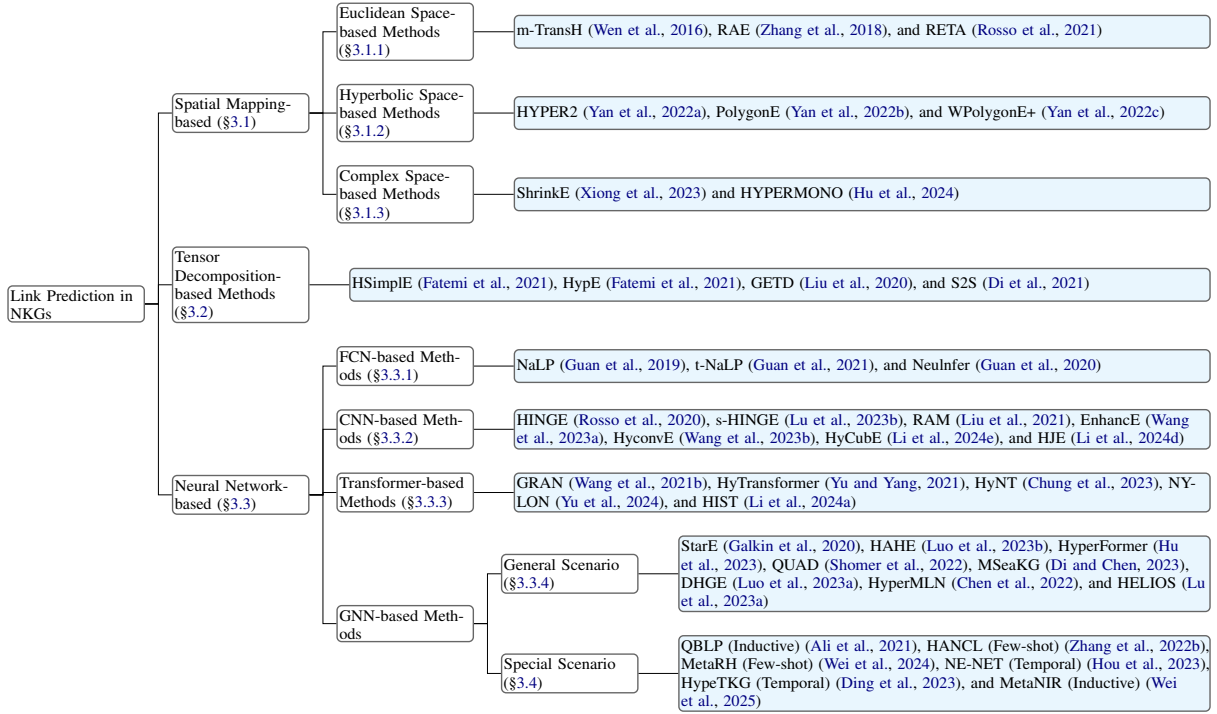
\begin{figure*}[htbp]
  \centering
  \resizebox{\textwidth}{!}{
    \begin{tikzpicture}
      \tikzset{
        grow'=right,level distance=25mm, sibling distance =3.5mm,
        execute at begin node=\strut,
    every tree node/.style={%red,
        draw=gray!80!black,
              line width=0.6pt,
              text width=2cm,
              rounded corners=2pt,
              minimum height=0.5cm,
              anchor = west,
              fill=white,
              minimum width=2mm,
              inner sep=1pt,
        align=left,
        font = {\scriptsize}},
           edge from parent/.style={draw=black,
           edge from parent fork right}
           }
           %%% =======================================================
           \begin{scope}[frontier/.style={sibling distance=4em,level distance = 7em}]
            \Tree
            [.{Link Prediction in NKGs}
            % [.{Boost LLMsPerformance(3)}
            [.{Spatial Mapping-based (§\hyperlink{smm}{3.1})}
                [.{Euclidean Space-based Methods (§\hyperlink{esm}{3.1.1})}
                \node[fill=CY,text width=10.5cm](t1){m-TransH~\cite{wen2016representation}, RAE~\cite{zhang2018scalable}, and RETA~\cite{rosso2021reta}};
                ]
                [.{Hyperbolic Space-based Methods (§\hyperlink{hsm}{3.1.2})}
                \node[fill=CY,text width=10.5cm](t1){HYPER2~\cite{yan2022hyper2}, PolygonE~\cite{yan2022polygone}, and WPolygonE+~\cite{yan2022modeling}};
                ]
                [.{Complex Space-based Methods (§\hyperlink{csm}{3.1.3})}
                \node[fill=CY,text width=10.5cm](t1){ShrinkE~\cite{xiong2023shrinking} and HYPERMONO~\cite{hu2024hypermono}}; %, BiVE~\cite{chung2023learning}, and NestE~\cite{xiong2023neste}};
                ]
                ]
                [.{Tensor Decomposition-based Methods (§\hyperlink{tdm}{3.2})}
                \node[fill=CY,text width=13cm](t1){HSimplE~\cite{fatemi2019knowledge}, HypE~\cite{fatemi2019knowledge}, GETD~\cite{liu2020generalizing}, and S2S~\cite{di2021searching}};
                ]
                [.{Neural Network-based (§\hyperlink{nnm}{3.3})}
                [.{FCN-based Methods (§\hyperlink{fcn}{3.3.1})}
                \node[fill=CY,text width=10.5cm](t1){NaLP~\cite{guan2019link}, t-NaLP~\cite{guan2021link}, and Neulnfer~\cite{guan2020neuinfer}};
                ]
                [.{CNN-based Methods (§\hyperlink{cnnm}{3.3.2})}
                \node[fill=CY,text width=10.5cm](t1){HINGE~\cite{rosso2020beyond}, s-HINGE~\cite{Lu2023schema}, RAM~\cite{liu2021role},  EnhancE~\cite{wang2023enhance}, HyconvE~\cite{wang2023hyconve}, HyCubE~\cite{li2024hycube}, and HJE~\cite{li2024hje}};
                ]
                [.{Transformer-based Methods (§\hyperlink{tm}{3.3.3})}
                \node[fill=CY,text width=10.5cm](t1){GRAN~\cite{wang2021link}, HyTransformer~\cite{yu2021improving}, HyNT~\cite{chung2023representation}, NYLON~\cite{yu2024robust}, and HIST~\cite{li2024integrating}};
                ]
                [.{GNN-based Methods}
                [.{General Scenario (§\hyperlink{gsm}{3.3.4})}
                \node[fill=CY,text width=8cm](t1){StarE~\cite{galkin2020message}, HAHE~\cite{luo2023hahe}, HyperFormer~\cite{hu2023hyperformer}, QUAD~\cite{shomer2022learning}, MSeaKG~\cite{di2023message}, DHGE~\cite{luo2023dhge}, HyperMLN~\cite{chen2022explainable}, and HELIOS~\cite{lu2023helios}};
                ]
                [.{Special Scenario (§\hyperlink{ssm}{3.4})}
                \node[fill=CY,text width=8cm](t1){QBLP (Inductive)~\cite{ali2021improving}, HANCL (Few-shot)~\cite{zhang2022true}, MetaRH (Few-shot)~\cite{wei2023few}, NE-NET (Temporal)~\cite{hou2023temporal}, HypeTKG (Temporal)~\cite{ding2023exploring}, and MetaNIR (Inductive)~\cite{wei2025inductive}};
                ]
                ]
                ]
                ]
    \end{scope}
  \end{tikzpicture}
  }
  \caption{Classification of Methods for Link Prediction in NKGs. “()” marks specific special settings.}
  \label{class_methods}
\end{figure*}

  \subsection{\textbf{Classification of Methods for Link Prediction in NKGs}}
  From a technical perspective, link prediction methods for NKGs fall into three main categories: spatial mapping-based, tensor decomposition-based, and neural network-based. Spatial mapping-based methods project entities into semantic space (e.g., Euclidean, hyperbolic, or complex) and then assess fact plausibility via entity positions. Tensor decomposition-based methods model n-ary facts as higher-order tensors, indicating fact validity. Neural network-based methods use Fully Connected Network (FCN), Convolutional Neural Network (CNN), Transformer, or Graph Neural Network (GNN) to encode element associations in n-ary facts.
  
  Most of the above methods are designed for the general scenario, while some GNN-based methods address special scenarios, such as the temporal, few-shot, and inductive settings. In addition, different methods use different formalizations of n-ary facts. The correspondence between fact formalizations and methods is shown in Appendix~\ref{formalization}.

  \subsection{Applicability of NKGs and Their Link Prediction}
  This subsection discusses the scenarios where NKGs are particularly beneficial and highlights the fundamental differences between link prediction in NKGs and traditional KGs.

  \subsubsection{When to Use NKGs: Suitable Scenarios and Motivations}
  NKGs extend traditional KGs by effectively representing facts that involve three or more entities. Their use is particularly advantageous in the following scenarios: (1) Multi-party participation: a fact involves multiple semantically related entities; (2) Semantic coupling: entities in facts form an inseparable semantic unit that cannot be split into multiple triples without loss of meaning; (3) Context-dependent facts: facts are context-dependent, requiring time, location, or other conditions to be fully understood. Assessing these aspects can help determine whether NKGs provide advantages over traditional KGs for more accurate downstream reasoning. Further details are provided in Appendix~\ref{applicability}.
  
  \subsubsection{Link Prediction in NKGs vs. Traditional KGs: Key Differences}
  While traditional KGs and NKGs share basic components (entities, relations, facts) and adopt similar techniques (e.g., spatial mapping, tensor decomposition, neural network methods), key differences in fact structure and task definition necessitate tailored modeling strategies for NKGs.
  
\textbf{Modeling Structure:} Traditional KG models handle simple triples $(h, r, t)$, whereas NKGs must represent complex n-ary facts, such as hyper-relational facts involving main triples and multiple qualifier role-value pairs, demanding higher expressive power.

\textbf{Prediction Tasks:} Traditional methods focus on completing missing elements in triples (e.g., $(h, r, ?)$), while link prediction in NKGs often involves multiple missing roles, values, or entities, requiring more flexible and robust reasoning.

These distinctions drive specific adaptations in NKG methods. See Appendix~\ref{compare_kg_nkg} for a detailed comparison.

  \section{\textbf{Methods for Link Prediction in NKGs}}\label{section3}

  This section begins with introducing link prediction approaches for NKGs in general scenarios, after which it examines methods tailored to specific scenarios such as temporal, inductive, and few-shot settings. For the methods in general scenarios, we introduce each category of methods one by one, first introducing their general ideas and then delving into specific methods.

  \subsection{\textbf{Spatial Mapping-based Methods}}\label{smm}
  These methods map entities into a shared embedding space, enforcing geometric constraints to ensure meaningful spatial relationships among them.
  They can be further divided into three types based on the embedding space: Euclidean, hyperbolic, and complex space-based methods.
  
  \textbf{Euclidean Space-based Methods.}\label{esm}
m-TransH~\cite{wen2016representation} projects entities in a fact onto a hyperplane according to their corresponding roles, and then evaluates the fact with spatial positions of the entities. However, its complexity grows with the number of missing entities in the n-ary fact. RAE~\cite{zhang2018scalable} reduces this complexity by assuming high similarity among entities within a fact and only calculating entities with high similarity.

\textbf{Hyperbolic Space-based Methods.}\label{hsm}
The number of related entities grows exponentially along the NKG hierarchy, which aligns with the superlinear growth in hyperbolic space. To capture such hierarchical structures, HYPER2~\cite{yan2022hyper2} and PolygonE~\cite{yan2022polygone} embed entities in hyperbolic space. HYPER2 projects entities to the tangent space to integrate qualifier values, then maps them back for scoring. PolygonE treats n-ary facts as gyro-polygons and evaluates entity compatibility via vertex-gyrocentroid distances, preserving both structure and semantics. To address the assumption of equal entity importance, WPolygonE+~\cite{yan2022modeling} introduces learned entity weights and enhances fact representation by linking gyro-midpoints and centroids.

\textbf{Complex Space-based Methods.}\label{csm}
These methods effectively capture inference patterns in complex space, particularly monotonicity: if two role-value pairs $q_i$ and $q_j$ satisfy $q_i$ implies $q_j$, then forattaching either to a fact should yield the same truth value. ShrinkE~\cite{xiong2023shrinking} models a primary triple as a spatial-functional transformation specific to its relation, mapping the head entity to a query box in complex space that contains potential answer entities. Each qualifier constrains this box by shrinking it, ensuring that the contracted box remains inside the original—providing a geometric view of monotonicity through box containment. HYPERMONO~\cite{hu2024hypermono} first aggregates neighbor information to enhance entity representation, and then to achieve qualifier monotonicity HYPERMONO adopts cone embedding. Each time a qualifier is added, the angle of the cone is reduced, thereby reducing the answer set.
  
  \subsection{\textbf{Tensor Decomposition-based Methods}} \label{tdm}
  Such methods represent the set of facts in an NKG as a high-order tensor, where each tensor entry indicates the truth value of a particular fact. By reconstructing, decomposing, and optimizing this tensor, the model uncovers latent pattern features among the elements of n-ary facts, thereby enhancing link prediction accuracy.
  HSimplE~\cite{fatemi2019knowledge} shifts entity embeddings by position and combines them with hyperedge embeddings for scoring. HypE~\cite{fatemi2019knowledge} enhances this by using convolutional filters to generate position-specific embeddings.

GETD~\cite{liu2020generalizing} generalizes Tucker decomposition by reshaping it into a higher-order tensor and applying tensor ring decomposition~\cite{wang2018wide} to reduce parameters. However, GETD cannot handle multiple facts of different number of entities at the same time, which easily leads to data sparsity problems. S2S~\cite{di2021searching} addresses this issue by partitioning embeddings to enable parameter sharing across facts of varying sizes, thereby improving efficiency.

  \subsection{\textbf{Neural Network-based Methods}} \label{nnm}
  These methods leverage neural networks to encode NKGs and perform link prediction with learned element representations.  They can be categorized into four types: FCN-based, CNN-based, Transformer-based, and GNN-based methods. Each type employs a corresponding neural network architecture to encode n-ary facts.
  
  \subsubsection{FCN-based Methods}\label{fcn}
  NaLP~\cite{guan2019link} models n-ary facts as sets of role-value pairs, using FCNs to extract and aggregate features for truth value prediction, but treats all pairs equally. To address this, NeuInfer~\cite{guan2020neuinfer} decomposes a fact into a primary triple and qualifiers, scoring both the triple and its compatibility with qualifiers via FCNs. t-NaLP~\cite{guan2021link} further enhances NaLP by incorporating entity types and improved negative sampling.
  
  \subsubsection{CNN-based Methods}\label{cnnm}
  HINGE extends NeuInfer by modeling n-ary facts as a primary triple with role-value pairs, using CNNs and min-pooling for feature aggregation and an FCN for scoring. s-HINGE~\cite{Lu2023schema} further incorporates entity type information to enhance performance, similar to t-NaLP.
  RAM~\cite{liu2021role} introduces a latent space to model role semantics, where role embeddings are generated via linear combinations of basis vectors, and role-specific pattern matrices evaluate entity-role compatibility using a multilinear scoring function.

To better exploit entity context, EnhancE~\cite{wang2023enhance} enriches entity representations with position and neighbor information, and integrates semantics into relation embeddings.
To leverage CNNs’ representational power, HyconvE~\cite{wang2023hyconve} uses 3D convolution with role-aware and position-aware filters to capture intricate intra-fact interactions. HJE~\cite{li2024hje} enhances HyconvE with learnable position embeddings, while HyCubE~\cite{li2024hycube} improves efficiency by introducing 3D circular convolutions and a masked stacking strategy.
  
  \subsubsection{Transformer-based Methods}\label{tm}
  HyTransformer~\cite{yu2021improving} utilizes a Transformer~\cite{vaswani2017attention} with a regularization layer to encode n-ary facts. It initializes position embeddings to represent type information of elements within n-ary facts, but does not explicitly model the relationship type between two elements in an n-ary fact. GRAN~\cite{wang2021link} addresses this by representing each n-ary fact as a heterogeneous graph and introducing multiple edge types to encode relationship types between elements. It processes these heterogeneous graphs using multiple fully connected attention layers with edge-aware biases, improving the performance of the model. 
  
  The above methods can only handle discrete entities, however, real-world NKGs often contain numeric entities. For example, a number-related qualifier role-value pair (starting time, 1911) is associated with a triple (J.R.R., educated at, Oxford University). Recognizing the importance of numeric values,  HyNT~\cite{chung2023representation} encodes numeric literals within both primary triples and qualifier role-value pairs with a context transformer and a prediction transformer.
  
  Different from the above methods, HIST~\cite{li2024integrating} integrates text information and structural information in NKG to enhance the representation of elements in NKGs. This method uses GNNs to extract structural information and effectively integrates text and structural information through structural soft prompt tuning~\cite{chen2023dipping}. Recently, NYLON~\cite{yu2024robust} extends GRAN to handle noisy NKGs, exploring robust link prediction in noisy NKGs. NYLON employs a Transformer with learnable edge biases to compute fact confidence and element confidence. Based on these confidences, efficient selective annotation is performed for annotation groups.
  
  \subsubsection{GNN-based Methods}\label{gsm}
  GNN-based methods for link prediction in NKGs in general scenarios fall into two categories: fact modeling and schema modeling.
  
\textbf{Fact Modeling Methods} focus on intra-fact message passing. StarE~\cite{galkin2020message} aggregates qualifier information into relations to update entity embeddings, but lacks reverse information flow. QUAD~\cite{shomer2022learning} improves this by enabling bidirectional aggregation between primary triples and qualifier pairs.
HAHE~\cite{luo2023hahe} models both global hypergraph structures and local semantic sequences using dual attention modules. HyperFormer~\cite{hu2023hyperformer} reduces noise from multi-hop neighbors via a bidirectional interaction mechanism and Mixture-of-Experts for parameter efficiency.
To address the rigidity of fixed architectures, MSeaKG~\cite{di2023message} introduces a neural architecture search framework with diverse message functions adaptable to various NKG formats.

\textbf{Schema Modeling Methods} enrich fact representations with schema information. DHKG~\cite{luo2023dhge} uses a dual-view encoder to model instance and ontology views, while HELIOS~\cite{lu2023helios} enhances type representation using GATs and self-attention. HyperCL~\cite{lu2024hypercl} further introduces hierarchical ontologies and concept-aware contrastive learning to balance fact and schema influences, achieving improved prediction performance.
  
  \subsection{\textbf{Special Scenario Methods}}\label{special_methods}
  Recent advances have extended GNN-based methods to specialized settings, including temporal, inductive, and few-shot scenarios.

\subsubsection{Temporal Setting}
N-ary facts often include temporal information, yet many methods either ignore it or treat time as a generic role, blurring the distinction between relational and temporal semantics. Without explicit temporal modeling, models fail to capture ordering, duration, and the influence of historical patterns on future facts. NE-Net~\cite{hou2023temporal} addresses this by leveraging an entity-role encoder based on a GNN to capture precise entity evolution representations. HypeTKG~\cite{ding2023exploring} further considers the influence of time-invariant relations on temporal reasoning.

\subsubsection{Inductive Setting}
In real-world scenarios, new elements often emerge after the training phase, presenting a challenge for methods that struggle to handle unseen elements. To handle unseen elements emerging post-training, QBLP~\cite{ali2021improving} generates embeddings from auxiliary facts and text. MetaNIR~\cite{wei2025inductive} adopts meta-learning~\cite{finn2017model} to simulate inductive tasks and generate adaptive embeddings. HART~\cite{yin2025inductive} combines hypergraph GNNs and Transformers with a role-aware mechanism to mine complex subgraph semantics for inductive prediction.

\subsubsection{Few-shot Setting}
Both few-shot and inductive settings address unseen elements, but few-shot learning focuses on scenarios with very limited examples rather than none. To predict links involving sparse relations, HANCL~\cite{zhang2022true} leverages GNNs and attention mechanisms to enhance entity representations and match queries to limited support instances. MetaRH~\cite{wei2023few} applies meta-learning to refine relation representations and improve generalization in few-shot settings.

\subsection{Model Selection Guidance}
In practical scenarios, model selection should balance predictive accuracy, scalability, and domain requirements. For high accuracy and complex reasoning, GNN-based models like HAHE are recommended, especially when relational structure and role semantics are critical. When computational resources are limited, simpler neural models (e.g., NeuInfer or HINGE) offer a trade-off between performance and efficiency. Tensor decomposition-based models such as S2S are suitable for scenarios that require high structural interpretability. Spatial mapping-based embedding methods are generally not recommended unless extreme efficiency is needed. Due to space limitations, the comparison of each type of method is shown in Appendix~\ref{section_comparison}.  

  \section{\textbf{Performance of Existing Methods}}\label{experiments}
  This section presents the benchmarks, evaluation metrics, and the performance of existing methods. Due to space constraints, we briefly report results in general scenarios here; results in special scenarios are provided in Appendix~\ref{performance}.
  \subsection{\textbf{Benchmarks}}
  JF17K~\cite{wen2016representation}, WikiPeople~\cite{guan2019link}, and WD50K~\cite{galkin2020message} are the most commonly used benchmarks for evaluating link prediction methods for NKGs in general settings. Specifically, JF17K is derived from Freebase~\cite{bollacker2008freebase}, while WikiPeople and WD50K are based on Wikidata~\cite{vrandevcic2014wikidata}. Table~\ref{tab:benchmark_datasets} shows the basic statistics of these datasets, where \#X is the number of X in the dataset, E and R represent entities and roles, respectively, Arity represents the number of entities in an n-ary fact, that is, N is the proportion of n-ary facts, respectively.

  \begin{table}[h]
    \small
    \centering
    \caption{Benchmarks of link prediction in NKGs in general scenarios. Statistics are based on the original paper.}
    \label{tab:benchmark_datasets}
    \begin{tabular}{lccccc}
      \toprule[1pt]
    \textbf{Dataset} & \textbf{\#E} & \textbf{\#R} & \textbf{Arity} & \textbf{N} & \textbf{\#Facts} \\
    \midrule[0.5pt]
    JF17K & 28,645 & 501 & 2-6  & 45.9\% & 100,947 \\
    WikiPeople & 47,765 & 193 & 2-9  & 11.6\% & 382,229 \\
    WD50K & 47,155 & 531 & 2-67 & 13.6\% & 236,507 \\
    \bottomrule[1pt]
    \end{tabular}
  \end{table}

  \subsection{\textbf{Metrics}}
  To evaluate link prediction, the model ranks candidate answers by score, aiming to rank the correct answer as high as possible. The two most widely used metrics are Mean Reciprocal Rank (MRR) and Hits@K. MRR is the average inverse rank of correct answers, while Hits@K is the proportion of correct answers ranked within the top $K$ positions (e.g., $K=1,3,5,10$). Both range from 0 to 1, with higher values indicating better performance. 

  \subsection{\textbf{Results}}
  Table~\ref{tab:nkg_results} reports the results of several representative link prediction methods on JF17K, WikiPeople, and WD50K, with best scores in bold. Neural network-based methods consistently outperform others, with HAHE (GNN-based) achieving the best result across all datasets, demonstrating the strength of GNNs in capturing complex entity interactions. In contrast, spatial mapping-based methods (m-TransH, RAE) perform worst, suggesting that simple projections are inadequate for n-ary fact modeling. Tensor decomposition-based methods (HypE, S2S) show moderate performance, but still fall short compared to neural network-based methods.
  
  \begin{table*}[ht]
    \centering
    \caption{Link prediction results in NKGs in general scenarios. ``-'' indicates that the method was not evaluated on the corresponding dataset in the original paper. Data for HypE and S2S are from~\cite{di2021searching}, others from~\cite{luo2023hahe}.}
    \small
    \label{tab:nkg_results}
    \begin{tabular}{llcccccc}
      \toprule[1pt]
      \multirow{2}{*}{\textbf{Type}} & \multirow{2}{*}{\textbf{Method}} & \multicolumn{2}{c}{\textbf{JF17K}} & \multicolumn{2}{c}{\textbf{WikiPeople}} & \multicolumn{2}{c}{\textbf{WD50K}} \\
                  &                 & \textbf{MRR}  & \textbf{Hits@1} & \textbf{MRR}  & \textbf{Hits@1} & \textbf{MRR}  & \textbf{Hits@1} \\
                  \midrule[0.5pt]
    \multirow{2}{*}{spatial mapping-based} 
                  & m-TransH        & 0.102 & 0.069  & -     & -      & -     & -      \\
                  & RAE             & 0.310 & 0.219  & 0.172 & 0.102  & -     & -      \\
                  \midrule[0.5pt]
    \multirow{2}{*}{Tensor Decomposition-based} 
                  & HypE            & 0.494 & 0.408  & 0.292 & 0.162  & -     & -      \\
                  & S2S             & 0.528 & 0.457  & 0.372 & 0.277  & -     & -      \\
                  \midrule[0.5pt]
    \multirow{4}{*}{Neural Network-based} 
                  & HINGE           & 0.473 & 0.397  & 0.333 & 0.259  & -     & -      \\
                  & NeuInfer        & 0.517 & 0.436  & 0.350 & 0.282  & 0.232 & 0.164  \\
                  & GRAN            & 0.656 & 0.582  & 0.479 & 0.410  & 0.309 & 0.240  \\
                  & HAHE            & \textbf{0.668} & \textbf{0.597}  & \textbf{0.495} & \textbf{0.420}  & \textbf{0.402} & \textbf{0.327}  \\
                  \bottomrule[1pt]
    \end{tabular}
    \end{table*}

  \section{\textbf{Applications of Link Prediction in NKGs}}\label{section5}
  Due to its ability to represent complex semantic relationships among multiple entities, link prediction in NKGs has shown great potential in various domains, including biomedicine, recommender systems, and financial technology. In biomedicine, it enables the modeling of intricate relationships such as drug-target-disease interactions, supporting applications like drug repositioning and personalized treatment planning. In recommender systems, NKG captures rich contextual signals (e.g., time, location, device), offering fine-grained user behavior modeling and improving recommendation accuracy. In financial technology, NKG supports the structured representation of multi-role financial facts, facilitating tasks like risk inference and high-risk case detection. Detailed examples and case studies are provided in Appendix~\ref{application}.

  \section{\textbf{Future Prospects}}\label{section6}
  Link prediction in NKGs has made significant progress but remains a nascent field. Further research is needed in the following directions.
  
  \subsection{\textbf{Link Prediction in NKGs with LLMs}}
LLMs have shown strong capabilities in both natural language understanding and structured data processing, including KGs. Recent work leverages this by converting KG elements into text and employing LLMs for link prediction. For example, KG-LLaMA~\cite{yao2023exploring} frames triples as textual sequences and fine-tunes LLaMA~\cite{touvron2023llama} to learn KG knowledge directly. KICGPT~\cite{wei2024kicgpt} addresses the hallucination issue in LLMs via a hybrid reranking mechanism that uses LLMs to refine candidates from traditional models. Sehwag et al.\shortcite{sehwag2024context} treat link prediction as prompt-based question answering, incorporating structural features like entity neighbors. CTLP\cite{li2024zero} employs structural paths between head and tail entities to conduct zero-shot link prediction. Similarly, KoPA~\cite{zhang2024making} converts KG structures, such as local subgraphs and relational context, into extended prompts, allowing LLMs to implicitly learn structural patterns through language. MuKDC~\cite{li2024llm} addresses few-shot learning challenges by prompting LLMs to generate synthetic structured knowledge, effectively augmenting training with high-quality pseudo-samples.

  Despite these advances, to the best of our knowledge, LLMs have not been applied to link prediction in NKGs, likely due to two main challenges: (1) converting structured n-ary facts into formats compatible with LLMs, and (2) overcoming input length limitations that hinder simultaneous processing of all candidate entities. Addressing these challenges could open new avenues for LLM-based link prediction in NKGs.
  
  \subsection{\textbf{Link Prediction in NKGs in Special Scenarios}}
  Most link prediction methods for NKGs focus on general scenarios. Research on special scenarios—temporal, inductive, and few-shot—remains in its early stages, offering ample opportunities for further exploration. For example, existing methods for temporal scenarios~\cite{hou2023temporal} overlook the local structure of n-ary facts, and existing methods for few-shot scenarios~\cite{zhang2022true, wei2023few} require extensive few-shot tasks for training, which are difficult to construct in real-world applications. Additionally, real-world NKGs are dynamic, frequently incorporating new facts. It is crucial to develop methods for growing NKGs that can adaptively learn from new facts while retaining previously acquired knowledge.
  
  \subsection{\textbf{Expainable Link Prediction in NKGs}}
  To our knowledge, HyperMLN~\cite{chen2022explainable} is the only method that explicitly addresses explainability in link prediction in NKGs. It employs a random field model to capture dependencies among n-ary facts and improves interpretability by extracting predefined first-order logic rules (e.g., self-inverse, symmetric, subrelation). However, its focus on predefined rule types limits its ability to explain more complex relational combinations frequently observed in n-ary facts. 
  Explainability should move beyond rule extraction to encompass broader interpretable reasoning, such as causal attribution~\cite{jaimini2024causallp} and counterfactual analysis~\cite{zhao2022learning}. Future work should pursue more flexible and comprehensive approaches to improve the transparency and reliability of link prediction in NKGs.
  
  \section{\textbf{Conclusion}}
  Link prediction in NKGs has emerged as a significant research area. In this survey, we provided the first comprehensive overview of existing work in this field. We began by introducing the definitions of NKGs and link prediction tasks within them, followed by a classification of current methods based on underlying techniques and application scenarios. Subsequently, we reported the performance and applications of existing methods. Finally, we outlined several promising future directions for advancing link prediction in NKGs.

\section*{Limitations}
This study presents a comprehensive overview of recent advances in link prediction in NKGs, covering a wide range of modeling paradigms and application scenarios. However, the current version primarily focuses on high-level comparisons of different approaches, with limited discussion on practical aspects such as computational efficiency, scalability to large-scale NKGs, and robustness to noise. In addition, while the survey categorizes methods based on modeling techniques and special settings, it does not deeply analyze cross-cutting concerns such as interpretability and generalization to out-of-distribution data. Moreover, the evaluation is based on a few widely used benchmarks, which may not fully reflect the challenges present in real-world applications. Future work could incorporate more detailed empirical analyses and consider broader deployment factors to offer a more holistic assessment of link prediction methods for NKGs.

\section{Ethics Statement}
This paper presents a comprehensive review of existing methods and research progress on link prediction in NKGs. It does not involve any experiments with human subjects, personal data, or potentially harmful content. All datasets and methods discussed are publicly available and widely used in academic research. Therefore, this work does not pose any ethical concerns.

% \section*{Acknowledgements}
% This document has been adapted by Yue Zhang, Ryan Cotterell and Lea Frermann from the style files used for earlier ACL and NAACL proceedings, including those for 
% ACL 2020 by Steven Bethard, Ryan Cotterell and Rui Yan

\bibliography{anthology,custom}
\bibliographystyle{acl_natbib}

\appendix

\newpage
\section{Limitations of Decomposition-Based Representations for N-ary Facts}\label{decomposition}
To represent n-ary facts using binary relational formats, link prediction methods for traditional KGs often rely on decomposition strategies such as reification~\cite{brickley2014rdf} and star-to-clique (S2C)~\cite{wen2016representation} transformation. These techniques convert each n-ary fact into multiple triples, enabling the use of standard KG embedding models originally designed for binary relations.

Reification introduces a virtual auxiliary entity to represent the original n-ary fact, and connects this auxiliary node to each participating entity through distinct role-specific relations. For instance, the 3-ary fact \textit{meeting(UN, Geneva, 2023)} can be decomposed into the binary facts (m1, \textit{organizer}, UN), (m1, \textit{location}, Geneva), and (m1, \textit{year}, 2023), where \texttt{m1} is the reified fact node. This approach preserves the full semantic structure of the original fact but increases graph complexity by introducing additional nodes and relations.

In contrast, the S2C approach avoids introducing auxiliary entities by directly connecting all participating entities in a fully connected subgraph (clique), assigning a specific relation to each entity pair. Using the same example, S2C may generate triples like (UN, \textit{met\_in}, Geneva), (UN, \textit{met\_on}, 2023), and (Geneva, \textit{hosted\_in}, 2023). This transformation simplifies the graph by keeping only the original entities, but it may obscure the unified semantic context of the original fact, making it harder for models to infer higher-level relations across multiple roles.

Overall, both decomposition strategies involve trade-offs: reification preserves relational integrity at the cost of increased structural complexity, while S2C maintains simplicity but risks losing contextual semantics essential for accurate reasoning.

\section{Positioning Our Work: A Survey Comparison on Link Prediction in NKGs}\label{survey}
Table~\ref{tab:1} provides a comprehensive comparison between our survey and existing surveys on link prediction in KGs. As shown, most existing surveys primarily focus on general KGs, temporal KGs, sparse KGs, or multi-modal KGs, and few of them systematically address link prediction in NKGs. Notably, while Shen et al. (2022) and Guan et al. (2022) partially cover NKGs by introducing task definitions and a limited number of methods, they lack a comprehensive exploration of NKG-specific models, benchmarks, and applications. In contrast, our survey specifically targets NKGs, providing a thorough task definition, a systematic summary of nearly 50 methods, over 10 benchmarks, as well as discussions on applications and future directions. To the best of our knowledge, our survey is the first to comprehensively address the link prediction task in NKGs.

\begin{table*}[htb]
  \small
  \setlength{\tabcolsep}{3pt} % 缩小列间距
	\centering
	\caption{Comparison of Existing Surveys on Link Prediction in KGs with Our Survey. “-” indicates that the survey does not introduce methods for link prediction in NKGs.}
	% \vspace{-1mm}
	\makebox[\columnwidth][c]{
		\begin{tabular}{cccc}
			\toprule[1pt]
      \textbf{Surveys} & \textbf{Years} & \textbf{KG types} & \makecell[c]{\textbf{Contents related to} \\\textbf{link prediction in NKGs}} \\
			\midrule[0.5pt]
            \makecell[c]{\citet{wang2017knowledge}, \citet{nguyen2017survey}, \\~\citet{dai2020survey}, ~\citet{ji2021survey}, \\~\citet{zou2020survey}, ~\citet{chen2020knowledge}, \\~\citet{chen2020review}, \citet{chen2021knowledge}, \\~\citet{rossi2021knowledge}, and ~\citet{wang2021survey}} & Before 2021 & General KG & -  \\			\midrule[0.5pt]
      \citet{zamini2022review} and ~\citet{ye2022comprehensive} & 2022 & General KG & - \\
			\midrule[0.5pt]
      \citet{chen2022survey} & 2022 & Temporal KG & - \\
			\midrule[0.5pt]
      \citet{zhu2022multi} & 2022 & Multi-modal KG & -  \\
			\midrule[0.5pt]
      \citet{shen2022comprehensive} & 2022 & General KG, temporal KG, and NKG & \makecell[c]{Contain task definition, \\4 methods, and \\2 benchmarks}  \\
			\midrule[0.5pt]
      \citet{guan2022event} & 2022 & General KG and NKG & \makecell[c]{Contain task definition \\and 4 methods} \\
			\midrule[0.5pt]
      \citet{liang2022survey} & 2022 & General, sparse, and multi-modal KG & -  \\
			\midrule[0.5pt]
      \citet{ferrari2022comprehensive} and ~\citet{hubert2023beyond} & 2023 & General KG & - \\
			\midrule[0.5pt]
	\makecell[c]{\citet{zhang2022few}, ~\citet{braken2023evaluation}, \\~\citet{ma2023survey}, ~\citet{chen2023zero}, \\and ~\citet{chen2023generalizing}} & 2023 & Sparse KG & - \\			\midrule[0.5pt]
      \citet{jiang2023evolution} & 2023 & General, sparse, and temporal KG & - \\
			\midrule[0.5pt]
      \citet{cai2022temporal} and ~\citet{wang2023survey} & 2023 & Temporal KG & - \\
			\midrule[0.5pt]
      \citet{peng2023multi} and ~\citet{chen2023survey} & 2023 & Multi-modal KG & - \\
			\midrule[0.5pt]
      \makecell[c]{\citet{cao2024knowledge}, ~\citet{ge2024knowledge}, \\~\citet{pote2024survey}, and ~\citet{mengstructure}} & 2024 & General KG & - \\
			\midrule[0.5pt]
      Ours & 2024 & NKG & \makecell[c]{Contain task definition, \\nearly 50 methods, \\more than 10 benchmarks, \\applications, and \\future prospect}  \\
      \bottomrule[1pt]
		\end{tabular}
	}
	\label{tab:1}
\end{table*}

\section{Evaluation Criteria for Modeling Facts with NKGs}\label{applicability}
NKGs are a generalization and extension of traditional KGs, providing the advantage of more accurately representing complex facts involving multiple entities. When a fact involves three or more core entities, NKGs can be prioritized to better capture such complex structures. To further assess whether the knowledge in a given domain is particularly suitable for modeling with NKG, three evaluation dimensions can be considered:
\begin{enumerate}
  \item \textbf{Multi-party Participation:} When a fact involves three or more entities that are semantically related, NKG is recommended to be used. The more participating entities involved in a fact, the stronger its multi-party participation. 
  For example, the fact ``A, B, C, and D are university classmates'' has a stronger multi-party participation compared to ``A, B, and C are university classmates,'' thus making NKG modeling more preferable for maintaining the completeness of the fact.
    
    \item \textbf{Semantic Coupling:} If the entities within a fact are tightly semantically coupled and cannot be reasonably decomposed into independent binary relations without losing essential semantics, NKGs should be used. For example, in the fact ``Student A received scholarship D at school C in year B'', all elements collectively form an inseparable semantic whole. 
    Decomposing it into multiple binary facts, such as (Student A, studies at, School C) and (Student A, received, Scholarship D), would fail to accurately capture the original semantics.
    
    \item \textbf{Context Dependence:} This refers to facts whose validity depends on contextual conditions such as time, location, or state. These contextual elements are integral parts of the fact’s semantics. In such cases, NKG is recommended, ensuring the completeness of contextual information. For instance, in the fact ``Einstein received the Nobel Prize in Physics in 1921 in Switzerland,'' both ``1921'' and ``Switzerland'' are essential contextual components.
     Ignoring them during modeling would compromise the accuracy of the fact. Therefore, such context-dependent facts are better modeled uniformly using NKG.
  \end{enumerate}
  Analyzing the above evaluation dimensions can support making choices between NKGs and traditional KGs, thereby improving the accuracy of downstream reasoning tasks.

\section{\textbf{Fact Formalization}}\label{formalization}
From the perspective of fact formalization methods, as previously discussed, current link prediction methods for NKGs can be broadly categorized into three types: hyperedge-based, role-value pair-based, and hyper-relational-based. The correspondence between these methods and their adopted fact representations is summarized in Table~\ref{tab:representation}. Different formalization methods directly affect the design of link prediction models in NKGs. 

For instance, hyperedge-based methods emphasize modeling the overall structural relationships among multiple entities within a fact and are adept at capturing complex interactions among entities. Role-value pair-based methods focus on role-centered modeling, which is effective in capturing the semantic influence of different roles on entities. Hyper-relational-based methods introduce modeling of entity importance, enabling a more precise reflection of the varying contributions and roles of entities within a fact.

Among these, hyper-relational formalization is the most widely applied in existing research. This approach not only possesses strong capability in modeling n-ary facts but also maintains compatibility with traditional KG triple facts, providing good flexibility and generalization ability. Consequently, it demonstrates strong adaptability and transferability across various real-world tasks.

Overall, when applying link prediction methods for NKGs to specific scenarios, researchers can flexibly select the most suitable fact formalization approach by referring to the comparative analysis in Figure~\ref{class_methods} and Table~\ref{tab:representation}, considering factors such as task types, data characteristics, and reasoning requirements.

\begin{table*}[ht]
\centering
\small
\caption{Classification of link prediction methods for NKGs by Fact Formalization Approach}
\label{tab:representation}
\begin{tabular}{ccc}
\toprule[1pt]
\textbf{Fact Formalization} & \textbf{Methods} & \textbf{Advantages} \\ 
\midrule[0.5pt]
Hyperedge-based & \makecell[c]{m-TransH~\cite{wen2016representation}, RAE~\cite{zhang2018scalable}, \\m-SimplE~\cite{fatemi2019knowledge}, HypE~\cite{fatemi2019knowledge}, \\GETD~\cite{liu2020generalizing}, S2S~\cite{di2021searching}, \\RAM~\cite{liu2021role}, EnhancE~\cite{wang2023enhance}, \\HyconvE~\cite{wang2023hyconve}, HyCubE~\cite{li2024hycube}, \\and HJE~\cite{li2024hje}} & \makecell[c]{Good at modeling the \\overall structural \\relationships among \\multiple entities \\in n-ary facts.} \\ 
\midrule[0.5pt]
Role-value pair-based & \makecell[c]{PolygonE~\cite{yan2022polygone}, NaLP~\cite{guan2019link}, \\t-NaLP~\cite{guan2021link}, NE-NET~\cite{hou2023temporal}, \\and HypeTKG~\cite{hou2023temporal}} & \makecell[c]{Effectively captures \\the semantic roles \\of each entity within \\n-ary facts.} \\ 
\midrule[0.5pt]
Hyper-relational-based & \makecell[c]{HYPER2~\cite{yan2022hyper2}, WPolygonE+\cite{yan2022modeling}, \\HYPERMONO\cite{hu2024hypermono}, NeuInfer~\cite{guan2020neuinfer}, \\HINGE~\cite{rosso2020beyond}, s-HINGE~\cite{Lu2023schema}, \\GRAN~\cite{wang2021link}, HyTransformer~\cite{yu2021improving}, \\HyNT~\cite{chung2023representation}, HIST~\cite{wang2023n}, \\NYLON~\cite{wang2023n}, StarE~\cite{galkin2020message}, \\HAHE~\cite{luo2023hahe}, HyperFormer~\cite{hu2023hyperformer}, \\QUAD~\cite{shomer2022learning}, DHGE~\cite{luo2023dhge}, \\HELIOS~\cite{lu2023helios}, HyperCL~\cite{chen2022explainable}, \\HANCL~\cite{zhang2022true}, and MetaRH~\cite{wei2023few}} & \makecell[c]{Effectively distinguishes \\the importance \\differences among \\entities within \\n-ary facts.} \\ 
\bottomrule[1pt]
\end{tabular}
\end{table*}

\section{Detailed Comparison between Link Prediction in NKGs and Link Prediction in Traditional KGs}~\label{compare_kg_nkg}
Traditional KGs and NKGs share some commonalities in their basic components, such as entities, relations, and facts. Consequently, they both adopt similar technical approaches for link prediction, including spatial, tensor decomposition, and neural networks. However, significant differences exist in the structural characteristics of their modeling targets and the definitions of their prediction tasks. These differences necessitate specific extensions and optimizations in the structural design and modeling strategies of link prediction methods for NKGs. Specifically, the key differences between link prediction in traditional KGs and link prediction in NKGs are reflected in the following two aspects.

\subsection{Modeling Object Structure} 
Traditional KG focuses on modeling triple facts $(h, r, t)$, while NKG deals with more flexible and complex n-ary facts. For example, for hyper-relational facts in the form of $(h, r, t), \{r_i : v_i\}_{i=1}^n$, link prediction in NKGs requires the model to not only capture the relation $r$ between the head entity $h$ and the tail entity $t$, but also handle the correspondence between qualifier roles $r_i$ and qualifier values $v_i$, as well as the interactions between these role-value pairs $\{r_i : v_i\}_{i=1}^n$ and the main triple $(h, r, t)$. Such complex structures within n-ary facts impose higher requirements on the model's representation capability.
\subsection{Prediction Task Definition} 
In traditional KGs, link prediction mainly focuses on completing missing entities within triples, such as $(?, r, t)$, $(h, r, ?)$, or $(h, ?, t)$. In contrast, link prediction in NKGs is broader and more flexible, involving not only missing entities and relations but also missing qualifier roles and values, and often requires the simultaneous completion of multiple missing elements. This task setting demands that the model possesses stronger structural modeling capabilities and flexible reasoning mechanisms to handle various types of missing elements.

Based on these differences, link prediction methods for NKGs have been specifically improved to address the modeling challenges of NKGs.
For instance:
\begin{itemize}
    \item \textbf{spatial mapping-Based Methods:} Introduce role-specific spatial transformation functions to capture semantic differences of entities under different role contexts.
    \item \textbf{Tensor Decomposition-Based Methods:} Utilize tools such as Tucker decomposition to handle higher-dimensional and structurally complex tensors, and address challenges such as nested structures and variable numbers of entities.
    \item \textbf{Neural Network-Based Methods:} Emphasize the modeling of unique structures in NKG, such as hypergraphs formed by inter-fact relations or fully connected graphs composed of elements within a fact.
\end{itemize}

Furthermore, in response to more complex prediction tasks, most current link prediction methods for NKGs support the completion of arbitrary missing elements, and some methods can even predict multiple missing elements simultaneously. For example, the HAHE method based on neural networks employs an autoregressive encoder combined with a MASK mechanism to effectively support the prediction of multiple missing elements within n-ary facts.

\section{\textbf{Comparison of link prediction methods for NKGs}}\label{section_comparison}
To intuitively compare the characteristics of the three types of methods, Table~\ref{class_comparison} shows the modeling idea, advantages, and drawbacks of each type of method. The spatial mapping-based methods have high computational efficiency, a small number of parameters, and are good at encoding large NKGs. However, due to the simple modeling idea, it has a great disadvantage in dealing with complex relationship types, and the model effect is usually poor. The tensor decomposition-based methods have a strong model expression ability in theory because their basic idea is to fully model the information contained in the NKG. However, it usually has more parameters, which is not conducive to application in large-scale NKGs. The neural network-based methods have strong feature learning ability, but the model is weak in interpretability.

\begin{table*}[htb]
  \setlength{\tabcolsep}{3pt} % 缩小列间距
  \centering
  \footnotesize
  \caption{Comparison of link prediction methods for NKGs.}
  \makebox[\columnwidth][c]{
      \begin{tabular}{cccc}
          \toprule[1pt]
          \textbf{Types} & \textbf{Introduction} & \textbf{Advantages} & \textbf{Drawbacks} \\
    \midrule[0.5pt]
    \makecell[c]{Spatial \\Mapping-based} & \makecell[c]{They use spatial transformations \\to model relationships between \\entities and roles in an n-ary \\fact, ensuring that their \\embedding vectors maintain \\specific geometric constraints \\in the embedding space.} & \makecell[c]{They are characterized by \\low time complexity, minimal \\model parameters, and fast \\training. This efficiency allows \\these methods to handle extensive \\datasets without significant \\computational overhead.} & \makecell[c]{They have limited expressive \\power, cannot capture the complex \\interactions between entities \\and roles, and usually have \\poor prediction results.}\\
    \midrule[0.5pt]
    \makecell[c]{Tensor \\Decomposition-\\based} & \makecell[c]{They represent an NKG as a \\high-order tensor where each \\cell indicates the validity \\of a corresponding fact.} & \makecell[c]{They possess strong expressive \\power, enabling them to capture \\complex relationships and interactions \\within NKGs. This results in \\relatively better model performance.} & \makecell[c]{
      They have high time complexity, \\especially for n-ary facts with \\high arity.}\\
    \midrule[0.5pt]
\makecell[c]{Neural \\Network-based} & \makecell[c]{
They use neural networks to \\extract features from n-ary \\facts and then score them \\to predict missing elements \\in NKGs.} & \makecell[c]{They effectively extract \\complex features from n-ary \\facts; they usually have \\high prediction accuracy.} & \makecell[c]{The training process of these \\methods usually requires a large \\amount of data and a long \\training time; they have poor \\interpretability.} \\
  \bottomrule[1pt]
  \end{tabular}
  }
  \label{class_comparison}
\end{table*}

\section{\textbf{More Details of Performance of Existing Methods}}\label{performance}
\subsection{\textbf{Benchmarks}}
\subsubsection{Temporal Scenario}
Hou et al.~\cite{hou2023temporal} constructed two datasets specifically for link prediction in NKGs in temporal scenarios: NWIKI and NICE. The NWIKI dataset is derived from Wikidata and contains a large number of n-ary facts, providing a rich foundation for link prediction in temporal NKGs. To ensure effective model training, they filtered out low-frequency entities and retained only high-frequency ones during the data construction process. In contrast, the NICE dataset is based on ICEWS, where the temporal information is more prominent, making it particularly suitable for tasks that require modeling dynamically evolving facts over time. These two datasets provide an important experimental foundation in the temporal NKG domain and have promoted research on n-ary fact modeling methods in temporal scenarios.
  
  At the same time, Di et al.~\cite{ding2023exploring} extended the traditional temporal KGs into temporal NKGs by identifying qualifying role-value pairs from Wikidata within the existing Wikidata11k~\cite{nobre1986inner} and YAGO1830~\cite{han2020explainable} datasets. The resulting datasets are named Wiki-hy and YAGO-hy, respectively. Table~\ref{tab:temporal_datasets} summarizes the statistics of NWIKI, NICE, Wiki-hy, and YAGO-hy datasets, where Timestamps indicates the recorded time points of facts, and Time Interval refers to the minimum time interval between facts.
  
\begin{table*}[ht]
  \small
  \centering
  \caption{Baseline Datasets for Link Prediction in NKGs in Temporal Scenarios. All statistics are reported from the original papers. ``Timestamps'' indicates the number of recorded time points, and ``Time Interval'' refers to the minimum temporal resolution of the dataset. ``N'' denotes the proportion of n-ary facts in the dataset.}
  \label{tab:temporal_datasets}
  \begin{tabular}{lccccccc}
  \toprule[1pt]
  \textbf{Dataset} & \textbf{\#Entities} & \textbf{\#Timestamps} & \textbf{Time Interval} & \textbf{N (\%)} & \textbf{\#Train} & \textbf{\#Valid} & \textbf{\#Test} \\
  \midrule[0.5pt]
  NWIKI    & 17,481  & 205    & 1 year     & 81.9\%  & 108,397  & 14,370  & 15,591  \\
  NICE     & 10,860  & 4,017  & 24 hours   & 97.5\%  & 368,868  & 5,268   & 46,159  \\
  Wiki-hy  & 11,140  & 507    & 1 year     & 9.5\%   & 111,252  & 13,900  & 13,926  \\
  YAGO-hy  & 10,026  & 188    & 1 year     & 6.9\%   & 51,193   & 10,973  & 10,977  \\
  \bottomrule[1pt]
  \end{tabular}
  \end{table*}
  
  \begin{table*}[ht]
    \small
    \centering
    \caption{Baseline Datasets for Few-shot Link Prediction in NKGs. All statistics are reported from the original papers. ``E-q'' and ``R-q'' indicate the number of entities and roles involved in qualifying role-value pairs, reflecting the complexity of n-ary facts in the dataset. ``Tasks'' refers to the number of few-shot relation link prediction tasks.}
    \label{tab:fewshot_datasets}
    \begin{tabular}{lccccccc}
    \toprule[1pt]
    \textbf{Dataset} & \textbf{\#Entities} & \textbf{\#Relations} & \textbf{\#E-q} & \textbf{\#R-q} & \textbf{N (\%)} & \textbf{\#Tasks} & \textbf{\#Facts} \\
    \midrule[0.5pt]
    WikiAnimals  & 2,925,278  & 167  & 396,739  & 48  & 49.7\%  & 49   & 5,964,839  \\
    WikiCompanies & 30,781    & 164  & 27,511   & 127 & 18.1\%  & 125  & 1,128,040  \\
    WD50K-Few    & 47,156     & 532  & 5,460    & 45  & 13.6\%  & 126  & 236,507    \\
    F-WikiPeople & 40,529     & 237  & 4,663    & 75  & 9.0\%   & 30   & 319,140    \\
    F-JF17K      & 19,721     & 480  & 4,928    & 127 & 47.6\%  & 52   & 91,572     \\
    F-WD50K      & 43,802     & 697  & 10,242   & 85  & 13.1\%  & 118  & 379,653    \\
    \bottomrule[1pt]
    \end{tabular}
    \end{table*}
    
    \subsubsection{Few-shot Scenario}
    Zhang et al.~\cite{zhang2022true} constructed three datasets for link prediction in NKGs under few-shot scenarios: WikiAnimals, WikiCompanies, and WD50K-Few. These datasets aim to simulate the learning challenges of rare relations in real-world settings and evaluate models' generalization abilities under limited data conditions. Specifically, WikiAnimals and WikiCompanies were derived from Wikidata by extracting facts related to animals and companies, respectively, while WD50K-Few was curated from a subset of the WD50K dataset. These datasets provide important experimental benchmarks for research on few-shot link prediction in NKGs.
    
    Concurrently, Wei et al.~\cite{wei2023few} further developed the F-WikiPeople, F-JF17K, and F-WD50K datasets by extending WikiPeople, JF17K, and WD50K, respectively, into few-shot scenarios. These datasets cover various knowledge domains and exhibit distinctive data distributions and n-ary fact structures, further enriching the experimental foundation for few-shot link prediction in NKGs. Table~\ref{tab:fewshot_datasets} summarizes the key statistics of these benchmarks for few-shot link prediction in NKGs, where ``E-q'' and ``R-q'' respectively represent the number of entities and roles involved in qualifying role-value pairs, reflecting the complexity of n-ary facts in each dataset, and ``Tasks'' indicates the number of few-shot relation link prediction tasks.
    
    \subsubsection{Inductive Scenario}
    Ali et al.~\cite{ali2021improving} constructed a series of datasets for inductive link prediction in NKGs based on WD50K, including multiple datasets under different settings to evaluate model generalization in inductive scenarios. This subsection focuses on the representative datasets WD20K(25), WD20K(100) V1, and WD20K(100) V2; for more details on other datasets, please refer to the original paper. In the inductive setting, these datasets typically include entities with textual descriptions or additional inference graphs (containing supporting instances related to unseen entities) to assist in generating representations for unseen entities during the testing phase. Specifically, WD20K(25) only provides textual descriptions without inference graphs containing unseen entities, requiring the model to rely solely on text features to complete inductive link prediction tasks. In contrast, WD20K(100) V1 and WD20K(100) V2 provide both textual descriptions and inference graphs, enabling models to leverage structural information to infer representations of unseen entities. Moreover, WD20K(100) V1 offers larger training data compared to WD20K(100) V2, allowing models to learn richer features during training. The number in parentheses in the dataset names indicates the proportion of hyper-relational facts, reflecting the diversity of facts across datasets. Furthermore, Wei et al.~\cite{wei2023few} constructed the JF-Ext, WIKI-Ext, and WD-Ext datasets, based on extensions of JF17K, WikiPeople, and WD50K, respectively, providing additional high-quality benchmarks for inductive link prediction in NKGs. These datasets cover different knowledge domains and vary in data scale, proportion of hyper-relational facts, and richness of entity information, thus offering a more comprehensive experimental benchmark for future research. Table~\ref{tab:inductive-datasets} summarizes the statistics of these benchmarks for inductive link prediction in NKGs. The ``Inference'' column indicates the inference graph used during testing.

    \begin{table*}[ht]
    \small
    \centering
    \caption{Benchmark datasets for inductive link prediction in NKGs. Statistics are from the original papers.}
    \label{tab:inductive-datasets}
    \begin{tabular}{lcccccccc}
      \toprule[1pt]
    \multirow{2}{*}{\textbf{Dataset}} & \multicolumn{2}{c}{\textbf{Train}} & \multicolumn{2}{c}{\textbf{Validation}} & \multicolumn{2}{c}{\textbf{Test}} & \multicolumn{2}{c}{\textbf{Inference}} \\
    & \textbf{\#Facts} & \textbf{N} & \textbf{\#Facts} & \textbf{N} & \textbf{\#Facts} & \textbf{N} & \textbf{\#Facts} & \textbf{N} \\
    % \cmidrule(lr){2-3} \cmidrule(lr){4-5} \cmidrule(lr){6-7} \cmidrule(lr){8-9}
    \midrule[0.5pt]
    WD20K(25) & 39,819 & 30.0\% & 4,252 & 25.0\% & 3,453 & 22.0\% & 0 & 0 \\
    WD20K(100) V1 & 7,785 & 100.0\% & 295 & 100.0\% & 364 & 100.0\% & 2,667 & 100.0\% \\
    WD20K(100) V2 & 4,146 & 100.0\% & 538 & 100.0\% & 678 & 100.0\% & 4,274 & 100.0\% \\
    JF-Ext & 3,305 & 54.0\% & 1,061 & 30.0\% & 1,283 & 21.0\% & 5,012 & 28.1\% \\
    WIKI-Ext & 3,905 & 2.1\% & 6,480 & 2.5\% & 4,733 & 2.9\% & 4,880 & 6.6\% \\
    WD-Ext & 5,112 & 1.0\% & 2,610 & 3.6\% & 3,053 & 2.0\% & 3,382 & 6.1\% \\
    \bottomrule[1pt]
    \end{tabular}
    \end{table*}
\subsection{Metrics}\label{matrics}
During evaluation, the model assigns scores to all candidate answers and ranks them accordingly. A higher rank for the correct answer indicates better performance. Mean Reciprocal Rank (MRR) and Hits@K are the most commonly used evaluation metrics, which assess the model's link prediction capabilities from different perspectives.

  \subsubsection{Mean Reciprocal Rank (MRR)}
  Mean Reciprocal Rank (MRR) is primarily used to evaluate the ranking quality of the correct answer among the prediction results, reflecting the overall link prediction performance of the model. The calculation formula is as follows:
  \begin{equation}
  \mathrm{MRR} = \frac{1}{|Q|} \sum_{q \in Q} \frac{1}{\mathrm{rank}_q},
  \end{equation}
  where $Q$ denotes the set of queries, and $\mathrm{rank}_q$ is the rank of the correct answer in the sorted list for query $q$. The value of MRR ranges from 0 to 1, where a higher value indicates better prediction performance.

  \subsubsection{Hits@K}
  Hits@K calculates the proportion of queries where the correct answer is ranked within the top $K$, measuring the model's performance at different precision levels to meet specific application needs. The formula is defined as:
  \begin{equation}
  \mathrm{Hits@K} = \frac{|\{q \in Q : \mathrm{rank}_q \leq K\}|}{|Q|}.
  \end{equation}

  Similar to MRR, Hits@K ranges from 0 to 1, with higher values indicating that the model is more capable of ranking the correct answer in the top positions. Common values of $K$ include 1, 3, 5, and 10, corresponding to different requirements of prediction accuracy in various application scenarios.

\subsection{\textbf{Results}}
\subsubsection{Temporal Scenario}
NE-Net and HypeTKG are specifically designed for link prediction in NKGs in temporal scenarios. Table~\ref{tab:temporal_results} presents their experimental results on the temporal datasets NWIKI and Wiki-hy, along with several representative baseline models. CEN~\cite{li2022complex}, TiGRN~\cite{li2022tirgn}, and DE-SimplE~\cite{goel2020diachronic} are typical temporal link prediction methods for binary facts, capable of effectively modeling temporal information in KGs but without considering the qualifier role-value pairs in multi-fact settings. HINGE, RAM, and HyTransformer are representative methods for link prediction in NKGs but do not leverage temporal information in temporal NKGs. The results demonstrate that NE-Net and HypeTKG achieve the best performance on the two datasets, significantly outperforming both temporal binary fact modeling methods and non-temporal NKG modeling methods. For example, NE-Net achieves an MRR of 0.720 on the NWIKI dataset, considerably surpassing other methods, further verifying its effectiveness in temporal link prediction in NKGs. These results indicate that jointly modeling qualifier role-value pairs and temporal evolution can significantly enhance predictive capability in temporal link prediction in NKGs.

\begin{table*}[h]
\small
\centering
\caption{Link prediction results in the temporal scenario. The results on NWIKI and Wiki-hy are from~\cite{hou2023temporal},~\cite{ding2023exploring}.}
\label{tab:temporal_results}
\begin{tabular}{lcccccc}
  \toprule[1pt]
\multirow{2}{*}{\textbf{Method}} & \multicolumn{3}{c}{\textbf{NWIKI}} & \multicolumn{3}{c}{\textbf{Wiki-hy}} \\
 & \textbf{MRR} & \textbf{Hits@1} & \textbf{Hits@10} & \textbf{MRR} & \textbf{Hits@1} & \textbf{Hits@10} \\
 \midrule[0.5pt]
HINGE & 0.217 & 0.191 & 0.259 & 0.543 & 0.497 & 0.694 \\
HypE & 0.252 & 0.249 & 0.257 & 0.624 & 0.604 & 0.658 \\
CEN & 0.406 & 0.302 & 0.610 & - & - & - \\
DE-SimplE & 0.138 & 0.108 & 0.191 & 0.351 & 0.218 & 0.640 \\
TiGRN & 0.611 & 0.506 & 0.811 & - & - & - \\
NE-Net & \textbf{0.720} & \textbf{0.668} & \textbf{0.802} & - & - & - \\
HypeTKG & - & - & - & \textbf{0.693} & \textbf{0.642} & \textbf{0.792} \\
\bottomrule[1pt]
\end{tabular}
\end{table*}

\subsubsection{Few-shot Scenario}
HANCL and MetaRH are specifically designed for few-shot link prediction in NKGs. Table~\ref{tab:fewshot_results} presents their experimental results on the few-shot datasets WikiAnimals and F-WD50K. FSRL~\cite{zhang2020few}, FAAN~\cite{sheng2020adaptive}, and MetaR~\cite{chen2019meta} are representative few-shot link prediction methods for binary facts, capable of learning relational representations from limited data but neglecting qualifier role-value pairs in multi-fact settings. The results show that HANCL and MetaRH achieve the best performance on both datasets, significantly outperforming other methods. For instance, HANCL achieves an MRR of 0.318 on the WikiAnimals dataset, showing a substantial improvement over other approaches. These results suggest that traditional few-shot learning methods are insufficient for capturing knowledge in multi-fact settings, while incorporating qualifier role-value pairs can effectively enhance reasoning capability and significantly boost prediction performance.

\begin{table*}[h]
\small
\centering
\caption{Link prediction results in the few-shot scenario. The results on WikiAnimals and F-WD50K are from~\cite{zhang2022true},~\cite{wei2023few}.}
\label{tab:fewshot_results}
\begin{tabular}{lcccccc}
  \toprule[1pt]
\multirow{2}{*}{\textbf{Method}} & \multicolumn{3}{c}{\textbf{WikiAnimals}} & \multicolumn{3}{c}{\textbf{F-WD50K}} \\
 & \textbf{MRR} & \textbf{Hits@1} & \textbf{Hits@10} & \textbf{MRR} & \textbf{Hits@1} & \textbf{Hits@10} \\
 \midrule[0.5pt]
StarE & 0.265 & 0.233 & 0.215 & 0.102 & 0.057 & 0.177 \\
GRAN & 0.253 & 0.199 & 0.221 & 0.126 & 0.077 & 0.222 \\
FSRL & 0.236 & 0.201 & 0.230 & - & - & - \\
FAAN & 0.270 & 0.225 & 0.246 & 0.116 & 0.059 & 0.226 \\
MetaR & - & - & - & 0.108 & 0.064 & 0.183 \\
HANCL & \textbf{0.318} & \textbf{0.288} & \textbf{0.258} & - & - & - \\
MetaRH & - & - & - & \textbf{0.192} & \textbf{0.109} & \textbf{0.340} \\
\bottomrule[1pt]
\end{tabular}
\end{table*}

\subsubsection{Inductive Scenario}
QBLP, MetaNIR, and HART are specifically designed for inductive link prediction in NKGs. Table~\ref{tab:inductive_results} shows their experimental results on the WD20K(100) V1, WD20K(100) V2, and WD-Ext datasets. BLP~\cite{chen2019meta} is a representative inductive link prediction method in the KG domain, which generates embeddings for unseen entities by encoding textual descriptions, while StarE and CompGCN~\cite{vashishth2019composition} generate embeddings for unseen entities based on neighborhood information in reasoning graphs. Both BLP and CompGCN overlook qualifier role-value pairs in multi-fact settings. The results show that HART and MetaNIR achieve competitive results, demonstrating that leveraging qualifier role-value pairs in multi-fact settings is essential for inductive link prediction in NKGs.

\begin{table*}[h]
\small
\centering
\caption{Link prediction results in the inductive scenario. The results on WD20K(100) V1 and V2 are from~\cite{yin2025inductive}, and WD-Ext results are from~\cite{wei2025inductive}.}
\label{tab:inductive_results}
\begin{tabular}{lccccccccc}
  \toprule[1pt]
\multirow{2}{*}{\textbf{Method}} & \multicolumn{3}{c}{\textbf{WD20K(100) V1}} & \multicolumn{3}{c}{\textbf{WD20K(100) V2}} & \multicolumn{3}{c}{\textbf{WD-Ext}} \\
 & \textbf{MRR} & \textbf{Hits@1} & \textbf{Hits@10} & \textbf{MRR} & \textbf{Hits@1} & \textbf{Hits@10} & \textbf{MRR} & \textbf{Hits@1} & \textbf{Hits@10} \\
 \midrule[0.5pt]
BLP & 0.057 & 0.019 & 0.123 & 0.039 & 0.014 & 0.092 & - & - & - \\
CompGCN & 0.104 & 0.057 & 0.183 & 0.025 & 0.007 & 0.053 & - & - & - \\
StarE & 0.112 & 0.061 & 0.212 & 0.049 & 0.019 & 0.110 & 0.079 & 0.021 & 0.131 \\
QBLP & 0.107 & 0.039 & 0.245 & 0.066 & 0.034 & 0.120 & - & - & - \\
HART & \textbf{0.385} & \textbf{0.294} & \textbf{0.522} & \textbf{0.258} & \textbf{0.176} & \textbf{0.468} & - & - & - \\
MetaNIR & - & - & - & - & - & - & \textbf{0.582} & \textbf{0.433} & \textbf{0.901} \\
\bottomrule[1pt]
\end{tabular}
\end{table*}

\section{Details Applications of Link Prediction in NKGs}\label{application}
Due to their ability to represent complex semantic relationships among multiple entities, NKGs have been widely adopted in knowledge modeling and reasoning tasks across various domains, including biomedicine, recommender systems, and financial technology. This section provides a detailed discussion of how link prediction in NKGs is applied in these representative scenarios and highlights its practical value.

\subsection{\textbf{Biomedicine}}

In the biomedical domain, knowledge often involves complex multi-entity relationships, such as ``a drug treats a disease by targeting a specific biomarker'' or ``a gene mutation causes a disease within a certain population.'' These facts require modeling of tightly connected semantic entities.

NKGs enable more expressive and accurate knowledge representation by explicitly specifying the roles of each entity (e.g., drug, target, disease, population). On this basis, link prediction in NKGs facilitates the discovery of novel associations, such as between drug combinations and indications, thus supporting tasks like drug repositioning and the identification of combination therapies. For instance, Wang et al.~\cite{wang2024mhre} constructed the MedCKG dataset from clinical data of China Medical University and applied link prediction in NKGs to assist in generating personalized treatment plans.

\subsection{\textbf{Recommender Systems}}

User behaviors in recommender systems are influenced by various contextual factors, such as time, location, device, and behavior type. A typical user action can be described as ``a user browses a product at a certain time, in a specific location, using a certain device.'' However, traditional KGs often model such behaviors as simple triples (e.g., (user, buy, item)), which fail to capture these contextual dependencies.

NKGs offer a more comprehensive representation by encoding multi-faceted facts (e.g., user, item, behavior type, time, location), enabling a finer-grained understanding of user behavior. Link prediction on NKGs can identify users' potential interests under specific contexts, thereby improving recommendation performance. For example, SDK~\cite{liu2023self} models such multi-entity interactions holistically and significantly improves the representation quality of users and items. Moreover, SDK demonstrates enhanced generalization in cold-start and sparse-data settings through multidimensional reasoning.

\subsection{\textbf{Financial Technology}}

In financial technology, real-world facts often involve multiple components, such as ``a bank issues a loan to a customer at a specific time for a particular project,'' or ``an institution invests in an asset in a certain market while facing specific risks.'' These complex relationships are difficult to model using traditional KGs.

NKGs allow for accurate representation of each component and its role (e.g., lender, borrower, purpose, risk level), enabling comprehensive modeling of financial interactions. This supports advanced analysis tasks such as risk assessment and retrieval of similar historical cases, greatly enhancing automation in financial data analytics. For example, Hou et al.~\cite{hou2023temporal} constructed a financial NKG based on real-world transaction data and applied link prediction in NKGs to identify potentially high-risk loan cases, significantly improving the effectiveness of risk warning systems.

\end{document}